\documentclass{article} 

\usepackage{preprint,times}

\iclrfinalcopy

\usepackage{wrapfig}
\usepackage{lineno}
\usepackage{subcaption}
\usepackage{listings}

\usepackage{amsmath,amsfonts,bm}

\def\eqref#1{equation~\ref{#1}}

\def\1{\bm{1}}

\DeclareMathAlphabet{\mathsfit}{\encodingdefault}{\sfdefault}{m}{sl}
\SetMathAlphabet{\mathsfit}{bold}{\encodingdefault}{\sfdefault}{bx}{n}

\usepackage{amsmath}
\usepackage{amssymb}
\usepackage{booktabs}
\usepackage{multirow}
\usepackage{makecell}
\usepackage{colortbl}
\usepackage{pifont}
\usepackage{graphicx}
\graphicspath{{figures/}} 

\usepackage{pifont}

\usepackage{bm}
\usepackage{algorithm}
\usepackage{algorithmic}

\usepackage{soul}

\usepackage{enumitem}

\usepackage{xspace}

\definecolor{Orchid}{RGB}{218,112,214} %
\definecolor{purple}{RGB}{230,70,151}
\definecolor{lightgray}{gray}{0.9} %

\definecolor{bad}{RGB}{220, 20, 60} 
\definecolor{good}{RGB}{34, 139, 34}

\usepackage{adjustbox}

\usepackage{caption}
\usepackage{subcaption}
\usepackage{graphicx}
\usepackage[dvipsnames]{xcolor} 
\definecolor{nicebg}{rgb}{0.98,0.98,0.98}
\usepackage{bbm}

\usepackage{tikz}
\usetikzlibrary{shapes,arrows,positioning,fit,backgrounds,shadows}
\usepackage{fontawesome5}
\usepackage{tablefootnote}

\usepackage{amsmath}
\usepackage{amssymb}
\usepackage{booktabs}
\usepackage{multirow}
\usepackage{makecell}
\usepackage[table]{xcolor}
\usepackage{graphicx}
\graphicspath{{figures/}} 

\usepackage{pifont}

\usepackage{enumitem}
\usepackage{bm}
\usepackage{algorithm}
\usepackage{algorithmic}
\usepackage{wrapfig}

\usepackage{soul}

\newcommand{\findingstylename}{finding}
\newcommand{\findingstyle}[1]{\renewcommand{\findingstylename}{#1}}

\newcommand{\findingstylefinding}[3]{%
    \begin{tcolorbox}[
        colback=white!90!gray,
        colframe=teal!60!black,
        arc=5pt,
        boxsep=5pt,
        left=10pt, right=10pt, top=2pt, bottom=2pt,
        boxrule=0.8pt,
        drop shadow=gray!50!white,
        enhanced jigsaw,
        before skip=8pt, after skip=8pt
    ]
    \noindent\faBookmark\hspace{0.5em}\textbf{Finding #1: #2} #3
    \end{tcolorbox}
}

\newcommand{\findingstyletakeaway}[3]{%
    \begin{tcolorbox}[
        colback=blue!3!white,
        colframe=blue!40!black,
        fonttitle=\bfseries,
        title={\faLightbulb\hspace{0.3em}Finding #1: #2},
        arc=2mm,
        boxrule=0.5pt,
        left=5pt, right=5pt, top=2pt, bottom=2pt,
        before skip=10pt, after skip=10pt
    ]
    #3
    \end{tcolorbox}
}

\newcommand{\finding}[3]{%
    \ifx\findingstylename\findingstylefindingname\findingstylefinding{#1}{#2}{#3}%
    \else\findingstyletakeaway{#1}{#2}{#3}\fi
}
\newcommand{\findingstylefindingname}{finding}

\definecolor{edgeblue}{RGB}{0, 0, 200}
\definecolor{edgegreen}{RGB}{0, 200, 0}
\definecolor{gptgreen}{RGB}{0, 166, 126}
\definecolor{scholarpurple}{RGB}{169, 1, 251}
\definecolor{bgcode}{rgb}{0.95,0.95,0.95}
\definecolor{githubgreen}{rgb}{0.564, 0.933, 0.564}
\definecolor{orange}{rgb}{1,0.5,0}

\definecolor{codegreen}{rgb}{0,0.6,0}
\definecolor{codegray}{rgb}{0.5,0.5,0.5}
\definecolor{backcolour}{RGB}{245,248,250}
\definecolor{emph}{RGB}{166,88,53}
\definecolor{nightblue}{RGB}{9,49,105}
\definecolor{keywords}{RGB}{207,33,46}
\definecolor{lightpurple}{RGB}{130,81,223}
\definecolor{examplebg}{RGB}{250,243,240}
\definecolor{codemph}{RGB}{150,30,30}

\newcommand{\Paragraph}[1]{\par\vspace{2pt}\noindent\textbf{#1}}

\newcommand{\styledquote}[3][scholarblue!8]{%
\begin{center}
\colorbox{#1}{%
    \hspace{0.1in}%
    \begin{minipage}{0.8\textwidth}
    \vspace{0.1in}
    \small\itshape
    #2
    \def\temp{#3}%
    \ifx\temp\empty
    \else
        \begin{flushright}
        \small\normalfont
        ---#3
        \end{flushright}
    \fi
    \vspace{0.1in}
    \end{minipage}%
    \hspace{0.1in}%
}
\end{center}
}

\makeatletter
\newcommand{\smallsym}[2]{#1{\mathpalette\make@small@sym{#2}}}
\newcommand{\make@small@sym}[2]{%
  \vcenter{\hbox{$\m@th\downgrade@style#1#2$}}%
}
\newcommand{\downgrade@style}[1]{%
  \ifx#1\displaystyle\scriptstyle\else
    \ifx#1\textstyle\scriptstyle\else
      \scriptscriptstyle
  \fi\fi
}
\makeatother

\usepackage{xcolor}

\definecolor{mscolor}{rgb}{0.1,0.1,0.9}

\newcommand{\note}[1]{\textcolor{mscolor}{\textit{[NOTE: #1]}}}

\newcommand{\nanogen}{\textsc{NanoGen}\xspace}
\newcommand{\simpleeval}{\textsc{SimpleEval}\xspace}
\newcommand{\diffbench}{\textsc{DiffusionBench}\xspace}

\usepackage{xspace} 

\usepackage{tcolorbox}
\usepackage{fontawesome5}
\tcbuselibrary{skins,breakable}

\newtcolorbox{takeaway}[1]{
    colback=blue!3!white,
    colframe=blue!40!black,
    fonttitle=\bfseries,
    title={\faLightbulb\hspace{0.3em}Takeaway #1},
    arc=2mm,
    boxrule=0.5pt,
    left=5pt,
    right=5pt,
    top=2pt,
    bottom=2pt,
    before skip=10pt,
    after skip=10pt
}

\newtcolorbox{openquestion}[1]{
    colback=orange!3!white,
    colframe=orange!60!black,
    fonttitle=\bfseries,
    title={\faQuestionCircle\hspace{0.3em}Open Research Question #1},
    arc=2mm,
    boxrule=0.5pt,
    left=5pt,
    right=5pt,
    top=2pt,
    bottom=2pt,
    before skip=10pt,
    after skip=10pt
}

\newcommand{\huggingface}{\raisebox{-1.5pt}{\includegraphics[height=1.05em]{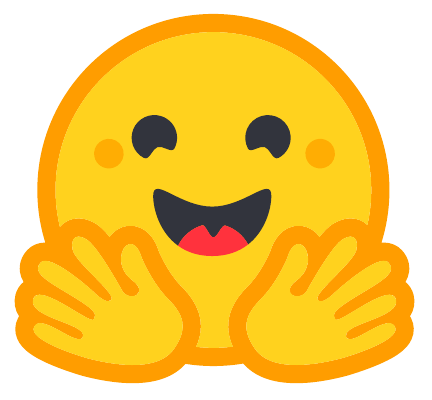}}\xspace}
\newcommand{\github}{\raisebox{-1.5pt}{\includegraphics[height=1.05em]{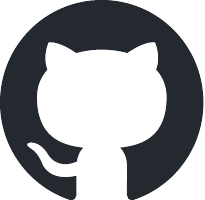}}\xspace}
\newcommand{\discord}{\raisebox{-1.5pt}{\includegraphics[height=1.05em,width=1.05em]{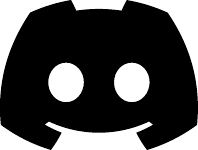}}\xspace}
\newcommand{\blogpost}{\raisebox{-1.5pt}{\includegraphics[height=1.05em]{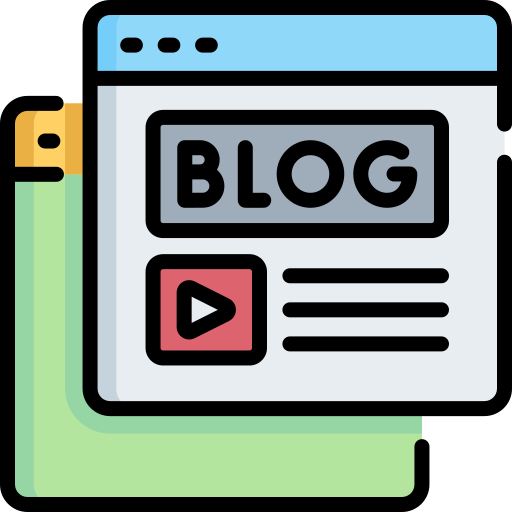}}\xspace}

\findingstyle{finding}

\usepackage[pagebackref=true]{hyperref}
\usepackage{url}

\definecolor{scholarblue}{rgb}{0.21,0.49,0.74}
\definecolor{darkblue}{rgb}{0.83, 0.89, 0.97}
\definecolor{cornellred}{rgb}{0.7, 0.11, 0.11}
\definecolor{cadmiumgreen}{rgb}{0.0, 0.42, 0.24}
\definecolor{aliceblue}{rgb}{0.91, 0.94, 0.97}
\definecolor{Red7}{rgb}{0.941, 0.243, 0.243}
\definecolor{Green7}{RGB}{55, 178, 77}
\definecolor{Blue9}{rgb}{0.28,0.5,0.95}

\hypersetup{
  breaklinks = true,
  linkcolor = cornellred,
  citecolor  = scholarblue,
  colorlinks = true,
  urlcolor = Blue9
}

\title{\raisebox{-7.5pt}{\includegraphics[height=1.5em]{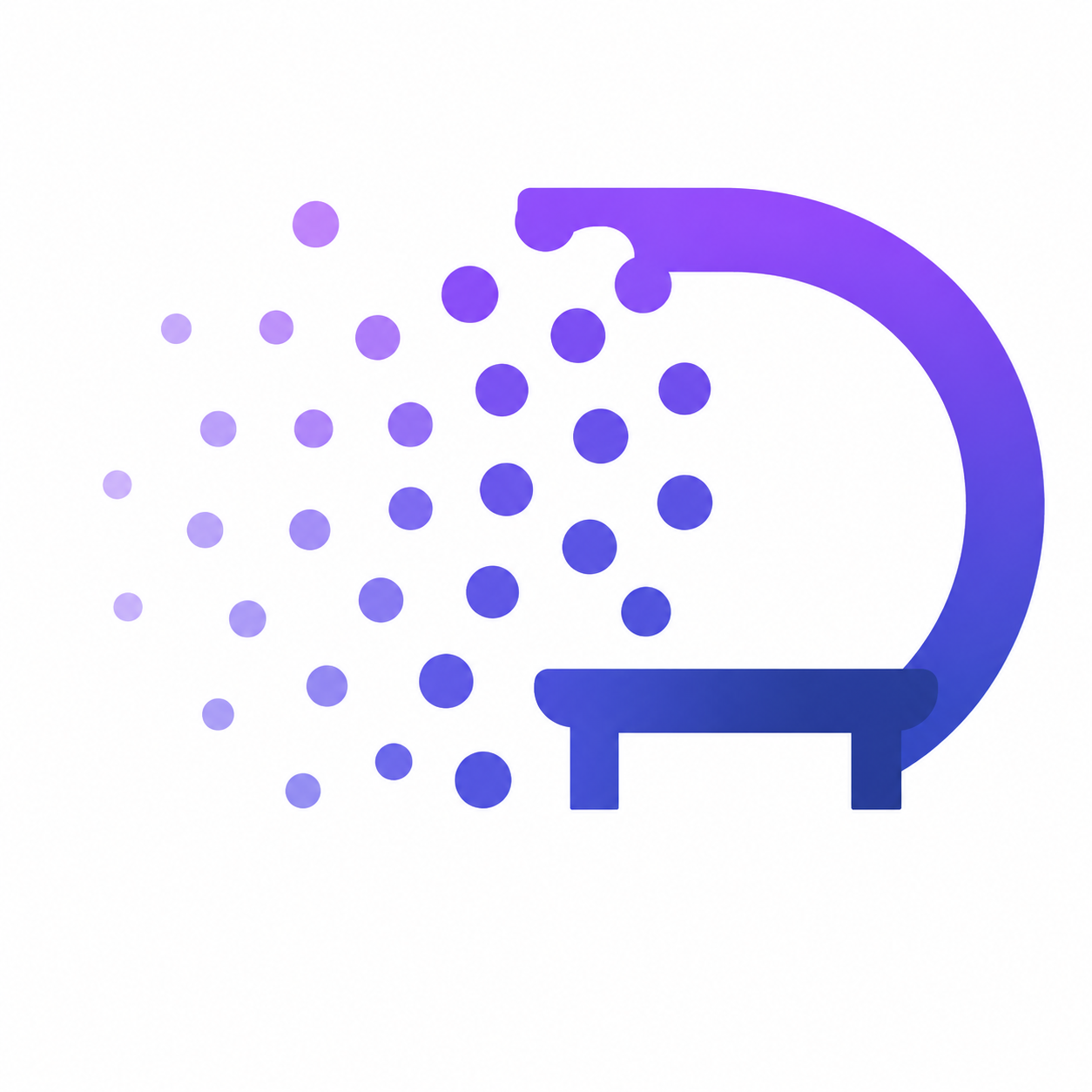}}\xspace
DiffusionBench: On Holistic Evaluation of \\Diffusion Transformers \\[0.4em]
{\normalsize\normalfont\itshape with a Unified Training Framework Bridging ImageNet and Text-to-Image}}

\newcommand{\anu}{$^{1}$}
\newcommand{\canva}{$^{2}$}

\author{Xingjian Leng{\anu\canva} \quad
  Jaskirat Singh{\anu} \quad
  Zhanhao Liang{\anu} \quad
  Ethan Smith{\canva}
  \\
  \bf Martin Bell{\canva} \quad
  Aninda Saha{\canva} \quad
  Yuhui Yuan{\canva} \quad
  Liang Zheng{\anu\canva}
  \\[0.3em]
  {\anu}Australian National University \quad
  {\canva}Canva Research
  \\
  \texttt{\{first-name.last-name\}@anu.edu.au} \\
  \texttt{\{ethansmith,martinbell,anindasaha,ryanyuan\}@canva.com}
}

\begin{document}

\maketitle

\begin{abstract}

Diffusion transformer (DiT) research on image generation has converged to a single evaluation setup: class-conditional generation on ImageNet. While methods improve the FID and related metrics, it is increasingly unclear whether they reflect real progress in generative modeling. The natural alternative, i.e., text-to-image (T2I) generation, is perceived as too costly or inconvenient to train and evaluate and is often skipped. We argue that this perception no longer holds. 
We introduce \nanogen, a unified DiT training and evaluation framework. \nanogen matches state-of-the-art DiT baselines on ImageNet and, with 12 lines of configuration change, also trains competitive text-to-image models. It currently supports RAE, VAE, pixel-space, and MeanFlow diffusion methods under both ImageNet and T2I setups. Under \nanogen, training T2I requires comparable compute to ImageNet. 
After training 21 latent diffusion models with \nanogen, we observe that method ranking shows no strong correlation between ImageNet and T2I generation: Pearson correlation is between -0.377 and -0.580 across three metrics. This suggests that a method which improves class-conditional ImageNet FID may show no corresponding improvement on T2I, clearly indicating the necessity of evaluating DiTs on both tasks. 
To this end, we summarize ImageNet and text-to-image results, which yields \diffbench, a holistic benchmark for DiT research. We recommend reporting \diffbench in place of ImageNet alone: methods that improve \diffbench are more likely to reflect broader progress. 

\begin{center}
    \urlstyle{same}
    \footnotesize
    \hspace*{-1.2cm}
    \begin{tabular}{rll}
        \github & \textbf{Code} & \url{https://github.com/End2End-Diffusion/diffusion-bench}\\
        \huggingface & \textbf{Models} & \url{https://huggingface.co/diffusion-bench}\\
        \discord & \textbf{Discord} & \url{https://discord.gg/jh5Bz8uHEr}\\
        \blogpost & \textbf{Blog} & \url{https://end2end-diffusion.github.io/diffusion-bench/}\\
    \end{tabular}
\end{center}

\end{abstract}



\section{Introduction}
\label{sec:intro}



Diffusion transformers (DiTs) have become the dominant method for image generation, with rapid progress over the last few years across architecture design~\citep{dit,sit,ddt,ldit}, training objectives~\citep{fm,jit,meanflow}, and representation learning~\citep{repa,sra,reg,irepa,repae}. Over the same period, the community has converged on a very narrow set of datasets for measuring this progress, \emph{e.g.}, most prominently, class-conditional ImageNet~\citep{imgnet} generation at 256 and 512 resolutions. The use of ImageNet-FID~\citep{fid} as a reporting standard has value: it makes method-method comparisons cheap and has accelerated progress on important modelling questions. 

\begin{figure*}[t]
  \centering
  \includegraphics[width=0.975\textwidth]{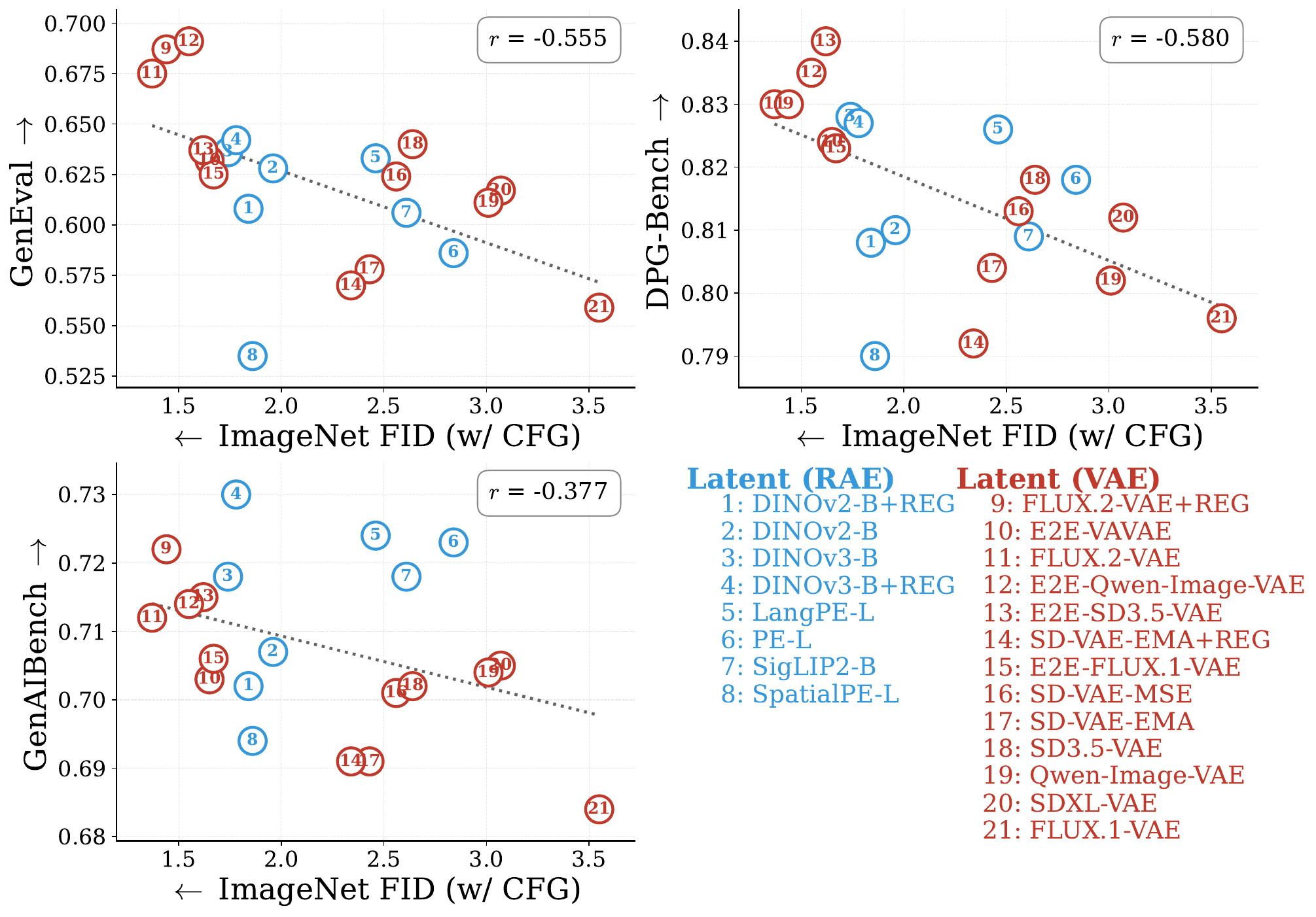}
  \caption{\textbf{Class-conditional ImageNet FID is not strongly correlated with T2I metrics for VAE and RAE methods.} Pearson correlations between ImageNet FID and a set of text-to-image evaluation metrics: GenEval~\citep{geneval}, DPG-Bench~\citep{dpgbench}, and GenAIBench~\citep{genaibench} across both RAE and VAE latent spaces. Results on ImageNet are evaluated under the best CFG scale of each method. We find no evidence of strong correlation across the three T2I metrics, indicating that ImageNet FID does not reliably predict T2I quality. We remove pixel-space methods from this comparison as they are far behind latent-space methods, artificially inflating the correlation. The corresponding without CFG version and the variants that additionally include three pixel-space methods are shown in Fig.~\ref{fig:fid_correlation_nocfg_appendix} and Fig.~\ref{fig:fid_correlation_withpixel_appendix} in the appendix.}
  \label{fig:fid_correlation}
\end{figure*}

However, 
with better modelling methods, it becomes increasingly hard to tell whether a reported gain on ImageNet indicates broadly better modelling or some kind of overfitting to the benchmark itself. The natural correction would be extending the evaluation to text-to-image (T2I) generation, a critical application of diffusion models, 
but in practice this does not happen very often. This is likely because training T2I models is perceived as high-cost and high-friction: they require different data pipelines, different evaluation procedures, and often an entirely different codebase. 

We challenge the premise that evaluating beyond ImageNet requires a separate and high-cost research programme. We introduce \nanogen, a DiT training framework whose ImageNet configuration matches state-of-the-art methods, and whose text-to-image configuration is reachable from the ImageNet configuration with roughly $12$ lines of config changes, covering the dataset and conditioning module.
Recent works such as i1~\citep{i1} and MiniT2I~\citep{minit2i} investigate simple recipes for training a strong T2I model, while our goal is to build a unified framework and compare the effectiveness of recent methods across tasks. With the shared backbone, optimiser, training loop, and evaluation suite, this paper asks a scientific question: if a method improves class-conditional ImageNet FID, does this imply a corresponding improvement on T2I generation?

For the frontier DiT models, we find there is no strong correlation between ImageNet FID and T2I performance. Using \nanogen, we train 21 latent diffusion models for ImageNet and T2I generation under nearly identical settings. As shown in Fig.~\ref{fig:fid_correlation}, class-conditional ImageNet-FID performance does not reliably predict T2I performance measured by metrics such as GenEval~\citep{geneval}, DPG-Bench~\citep{dpgbench}, and GenAIBench~\citep{genaibench}. As further shown in Fig.~\ref{fig:fid_correlation_nocfg_appendix} in the Appendix, this problem is consistent regardless if classifier-free guidance is applied. 
In other words, a technique that improves over existing methods on ImageNet generation may not exhibit such improvement on T2I generation. Without more holistic evaluation, progress under ImageNet may not generalize. 

Using \nanogen, we are able to obtain and combine ImageNet and T2I generation results using a range of metrics into a single benchmark, \diffbench. Since ImageNet rankings do not reliably predict T2I performance (Fig.~\ref{fig:fid_correlation}), our core recommendation is that future DiT work report \diffbench rather than ImageNet alone. Methods that improve \diffbench are then more likely to reflect broadly useful progress.


We highlight the main contributions of this paper below:
\begin{itemize}[leftmargin=1.4em,itemsep=2pt]
    \item We release \nanogen, a unified DiT training framework (Sec.~\ref{sec:method}) that matches state-of-the-art methods on ImageNet (Tab.~\ref{tab:in1k_reproducibility}) and extends to text-to-image training with roughly $12$ lines of config changes.
    \item We show empirically that ImageNet rankings do not reliably predict text-to-image performance (Fig.~\ref{fig:fid_correlation}, Tab.~\ref{tab:in1k_systematical_cfg} and Tab.~\ref{tab:t2i_systematical}), with magnitudes large enough to flip conclusions from ImageNet.
    \item We incorporate both tasks into \diffbench (Sec.~\ref{sec:diffbench}; Tab.~\ref{tab:in1k_systematical_cfg} and Tab.~\ref{tab:t2i_systematical}), present results of many existing methods, and suggest for its adoption as a default DiT benchmark.
\end{itemize}

\section{The \nanogen Training and Evaluation Framework}
\label{sec:method}


\nanogen is a diffusion model training and evaluation framework that supports both class-conditional ImageNet generation and text-to-image generation under a single codebase. Its goal is to make the additional cost of evaluating a method on the T2I task as low as possible, so that ``Did this idea also help on T2I?'' is a question any author can answer without re-inventing the wheel.

\subsection{Overall Architecture}
\Paragraph{Design principles.} \nanogen uses one DiT backbone, one optimiser, one training loop, one evaluation harness, and one config format for both ImageNet and T2I tasks. Switching between tasks requires only two changes: (i) the data pipeline is pointed at a different dataset: class-labelled ImageNet for class-conditional generation, or captioned images for T2I; (ii) the conditioning module is swapped accordingly: a class embedder for ImageNet, or a frozen text encoder for T2I. Everything else remains the same. As a result, moving \nanogen to a new task is equivalent to switching a dataset and a conditioner, rather than rewriting the stack. \nanogen supports many recent diffusion methods, including RAE~\citep{rae}, VAE~\citep{vae}, pixel-space, REG~\citep{reg}, and MeanFlow~\citep{meanflow} methods. \nanogen supports RAEv2~\citep{raev2} vision encoders and tokenizers.

\Paragraph{Backbone architecture.} We use a standard diffusion transformer with three deliberate modifications relative to the common DiT recipe.
\begin{itemize}[leftmargin=1.4em,itemsep=2pt]
    \item \underline{Decoupled Diffusion Transformer (DDT)}~\citep{ddt}. We use the DDT backbone as RAE~\citep{rae} does, splitting the model into an encoder and a decoder. The encoder takes the noisy input together with the conditioning tokens and produces a semantic representation. The decoder is a shallow but wide transformer, which takes that representation and the noisy entity as input and predicts the diffusion target. This split increases effective width without the quadratic FLOPs cost of a uniformly wide DiT.
    \item \underline{No AdaLN in the encoder.} Similar to iMeanFlow~\citep{imf}, i1~\citep{i1}, and MM-JiT~\citep{minit2i}, we remove the AdaLN modules from encoder blocks and retain AdaLN in the decoder. The modulation in the decoder is computed from the semantic output of the encoder rather than directly from the timestep.
    \item \underline{In-context conditioning.} We feed all conditioning, including the timestep, to the encoder as tokens prepended to the visual tokens. The encoder takes one token sequence as input and does not need task-specific modulation.
\end{itemize}
Because all conditioning is in-context, adding or removing conditioning for a new task only requires changing the conditioning tokens. The rest of the architecture is identical across tasks.

\Paragraph{Task-specific conditioning tokens.} The only per-task difference is the number and meaning of conditioning tokens:
\begin{itemize}[leftmargin=1.4em,itemsep=2pt]
    \item \underline{ImageNet (class-conditional).} 4 timestep tokens plus 8 class-conditioning tokens.
    \item \underline{Text-to-image.} 4 timestep tokens plus 256 text-conditioning tokens.
\end{itemize}
Everything else, such as backbone, loss formulation, optimiser, and EMA, is shared. This allows for moving a method to T2I generation with a small change rather than extensive engineering effort.

\Paragraph{Training recipe.} Wherever possible we keep optimisation-level choices fixed across tasks. We use the AdamW~\citep{adamw} optimiser with $\beta_1 = 0.9$ and $\beta_2 = 0.95$, and a learning rate that linearly warms up to $2{\times}10^{-4}$ and then linearly decays to $2{\times}10^{-5}$. We apply gradient clipping at $1.0$ and maintain an exponential moving average (EMA) of model weights with decay $0.9995$. For the diffusion schedule, training timesteps are sampled from a logit-normal distribution with mean $0$ and standard deviation $1$. Following SD3~\citep{sd3} and RAE~\citep{rae}, we additionally apply dimension-dependent timestep shifting $t_m = \frac{\alpha t_n}{(1 + (\alpha-1)t_n)}$, where $\alpha = \sqrt{\frac{m}{n}}$, $n = 4,096$, and $m$ is the effective input dimension. We use $v$-prediction by default. For methods whose original recipe specifies $x$-prediction, \textit{e.g.}, JiT~\citep{jit}, PixelGen~\citep{pixelgen}, we preserve $x$-prediction for faithful reproduction. For sampling, we default to an Euler sampler with 50 function evaluations (NFEs). On ImageNet we report results both with and without classifier-free guidance (CFG)~\citep{cfg}; on T2I we report only the with-CFG results. Data-dependent hyperparameters such as batch size, total training budget, and warmup duration also differ by task and are specified per section. For MeanFlow~\citep{meanflow} models, we use 75\% flow-matching loss mixed with 25\% MeanFlow loss. We set $\kappa = 0$ for the without-CFG experiments and $\kappa = 0.5$ for the with-CFG experiments.

\Paragraph{Evaluation protocols.} \nanogen supports unified online evaluation during training. The evaluation harness depends only on the HuggingFace Transformers library and requires no other packages. On ImageNet, we support FID~\citep{fid}, IS~\citep{is}, FDr~\citep{fdloss}, and MIND~\citep{mind}. On T2I, GenAIBench (VQAScore)~\citep{vqascore}, DPG-Bench~\citep{dpgbench}, and GenEval~\citep{geneval} are supported. Users can track training progress across all metrics in real time.

\subsection{ImageNet Generation in \nanogen}
\label{sec:imagenet}


Before evaluating on the text-to-image task, we first confirm that \nanogen establishes trustworthy ImageNet baselines, where the implemented methods match reported numbers in their papers. 

\textbf{Implementation details}. Using \nanogen, we train DiT models on ImageNet~\citep{imgnet} at $256 \times 256$ resolution for both latent-space and pixel-space generation. For data pre-processing, we follow ADM~\citep{adm} and center-crop images to $256 \times 256$. For the model backbone, we use a DDT with a 28-layer encoder of width 1,152 and a 2-layer decoder of width 2,048, yielding $\sim$615M parameters. We use a budget of 80 epochs at batch size 1,024 and 40 epochs of warmup for the learning-rate schedule. For evaluation, we follow the standard ImageNet protocol and generate 50,000 images, with 50 samples per class. We report FID~\citep{fid}, IS~\citep{is}, FDr~\citep{fdloss}, and MIND~\citep{mind} metrics.


\begin{table}[t]
    \centering
    \scriptsize
    \setlength{\tabcolsep}{4mm}{
    \begin{tabular}{lccccccc}
    \toprule
    \textbf{Method} & \textbf{Epochs} & \textbf{\#Params} & \textbf{Pred.} & \textbf{NFE} & \textbf{with Guidance} & \textbf{FID}$\downarrow$ & \textbf{IS}$\uparrow$ \\
    \midrule
    \multicolumn{8}{c}{\textbf{Latent-space}} \\
    \midrule
    RAE (DINOv2-B) & 80 & 839M & $v$ & 50 & $\times$ & 2.16 & 214.8 \\
    \rowcolor{blue!10} \textbf{Ours} & 80 & 847M & $v$ & 50 & $\times$ & \textbf{2.07} & \textbf{213.5} \\
    E2E-VAVAE & 80 & 675M & $v$ & 250 & $\times$ & 5.26 & - \\
    \rowcolor{blue!10} \textbf{Ours} & 80 & 680M & $v$ & 250 & $\times$ & \textbf{3.64} & \textbf{152.5} \\
    E2E-VAVAE + REPA & 80 & 675M & $v$ & 250 & $\times$ & 3.46 & 159.8 \\
    \rowcolor{blue!10} \textbf{Ours} & 80 & 681M & $v$ & 250 & $\times$ & \textbf{2.88} & \textbf{165.4} \\
    \midrule
    \multicolumn{8}{c}{\textbf{Pixel-space}} \\
    \midrule
    PixNerd & 160 & 458M & $v$ & 100 & \checkmark & 2.64 & 297.0 \\
    \rowcolor{blue!10} \textbf{Ours} & 160 & 446M & $v$ & 100 & \checkmark & \textbf{2.58} & \textbf{299.3} \\
    JiT & 200 & 131M & $x$ & 50 & \checkmark & 8.62 & - \\
    \rowcolor{blue!10} \textbf{Ours} & 200 & 88M & $x$ & 50 & \checkmark & \textbf{5.49} & \textbf{231.6} \\
    PixelGen & 40 & 459M & $x$ & 50 & $\times$ & 7.53 & 131.7 \\
    \rowcolor{blue!10} \textbf{Ours} & 40 & 458M & $x$ & 50 & $\times$ & \textbf{7.52} & \textbf{123.5} \\
    \bottomrule
    \end{tabular}
    }
    \caption{\textbf{ImageNet-256 reproducibility across latent-space and pixel-space methods.} For each method, we use AdamW and follow the original architecture. We build a model of similar size by adjusting the transformer width and depth, and train it for the same number of epochs as in the original paper. Evaluation follows the original setup of each method. Our results match or improve on published FID and IS with a unified codebase.}
    \label{tab:in1k_reproducibility}
\end{table}

\textbf{ImageNet reproducibility validation}. 
We implement and re-train six existing methods using \nanogen, including three latent-space methods: RAE~\citep{rae} and two E2E-VAEs~\citep{repae}, and three pixel-space methods: PixNerd~\citep{pixnerd}, JiT~\citep{jit}, and PixelGen~\citep{pixelgen}. We summarize their reported numbers and the \nanogen results in Tab.~\ref{tab:in1k_reproducibility}, where we try our best to keep major hyperparameters similar, \textit{e.g.}, model size, inference NFEs, and CFG settings. We observe that the \nanogen results are competitive with published numbers and sometimes slightly superior. This validates the effectiveness of \nanogen. We treat this as a necessary pre-condition for the cross-task analysis that follows, not as a contribution in itself.

\subsection{Text-to-Image Generation in \nanogen}
\label{sec:t2i}


Text-to-image generation is the second axis of DiT training and evaluation in \nanogen. Because ranking of methods measured on ImageNet does not appear consistent on T2I, as shown in Fig.~\ref{fig:fid_correlation}, it is important to evaluate method effectiveness not only on ImageNet, but also on T2I. 
This section first shows that moving from ImageNet training to T2I training is a small change in \nanogen. We then use existing T2I metrics, such as GenEval~\citep{geneval} and DPG-Bench~\citep{dpgbench}, to compare DiT methods in Sec.~\ref{sec:diffbench}.

\textbf{Implementation details}.
\label{sec:t2i_impl}
The T2I configuration of \nanogen is reached from the ImageNet configuration by replacing the class-embedding conditioner with a text-encoder conditioner and switching the dataset loader to a captioned image corpus. We use Qwen3-0.6B~\citep{qwen3} as the text encoder and take its final hidden states as the text-conditioning tokens. Pre-training uses the JourneyDB~\citep{journeydb}, Long-Caption, and Short-Caption splits of BLIP-3o~\citep{blip3o}. We use a batch size of $1024$ and $10\%$ conditioning dropout for CFG, run for $100\text{K}$ iterations. All other choices, including AdamW optimiser, learning-rate schedule, EMA, gradient clipping, $v$-prediction objective, and sampler, are inherited from the recipe in Sec.~\ref{sec:method}. For evaluation, we apply a classifier-free guidance scale of $6.0$ across the entire timestep interval. To avoid metric hacking on specific T2I benchmarks, we report results from the pre-training stage only and skip supervised fine-tuning on datasets like BLIP-3o-60K. 

\section{\diffbench: A Holistic Benchmark}
\label{sec:diffbench}
\subsection{Benchmarking ImageNet Generation Methods}
\textbf{Latent-space generation.}
In the latent-space regime we train \nanogen on both RAE~\citep{rae} and VAE latents. For RAE, we replace the latent encoder with six frozen pretrained vision encoders spanning two backbone scales and several pretraining objectives: ViT-B encoders DINOv2-B~\citep{dinov2}, DINOv3-B~\citep{dinov3}, and SigLIP2-B~\citep{siglip2}; and ViT-L encoders PE-L, LangPE-L, and SpatialPE-L from the Perception Encoder family~\citep{perceptionencoder}. For VAE, we train a range of widely used VAEs, including SD-VAE~\citep{ldm}, SDXL-VAE~\citep{sdxl}, SD3.5-VAE~\citep{sd3}, FLUX.1-VAE~\citep{flux}, FLUX.2-VAE~\citep{flux2}, Qwen-Image-VAE~\citep{qwenimage}, and VA-VAE~\citep{ldit}, together with their REPA-E~\citep{repae} end-to-end variants. Results are summarized in Tab.~\ref{tab:in1k_systematical_cfg}, evaluated with the best per-method CFG scale over the timestep interval $[0.0, 0.9]$. Note that our goal is not to claim a new state of the art on ImageNet. We have three main observations.

\begin{table*}[t]
\centering
\scriptsize
\setlength{\tabcolsep}{0.45mm}
\begin{tabular}{l|ccccc|ccccc|cc}
\toprule
\multirow{2}{*}{\textbf{Method}} & \multicolumn{5}{c|}{\textbf{FDr}$\downarrow$} & \multicolumn{5}{c|}{\textbf{MIND}$\downarrow$} & \multirow{2}{*}{\textbf{FID}$\downarrow$} & \multirow{2}{*}{\textbf{IS}$\uparrow$} \\
\cmidrule(lr){2-6} \cmidrule(lr){7-11}
 & \textbf{Incep.} & \textbf{ConvNeXt} & \textbf{DINOv2} & \textbf{MAE} & \textbf{SigLIP} & \textbf{Incep.} & \textbf{ConvNeXt} & \textbf{DINOv2} & \textbf{MAE} & \textbf{SigLIP} & & \\
\midrule
\multicolumn{13}{c}{\textbf{Latent-space (RAE)~\citep{rae}}} \\
\midrule
DINOv2-B & 1.22 & 2.20 & 3.26 & 6.19 & 7.76 & 2.03 & 66.49 & 27.74 & 0.43 & 6.52 & 1.96 & 224.1 \\
DINOv2-B + REG & 1.15 & 2.15 & 3.21 & 6.43 & 7.71 & 1.90 & 64.30 & 27.40 & 0.44 & 6.43 & 1.84 & 236.2 \\
DINOv3-B & 1.09 & 2.10 & 3.30 & 6.53 & 7.66 & 1.80 & 65.83 & 29.35 & 0.43 & 6.03 & 1.74 & 244.2 \\
DINOv3-B + REG & 1.11 & 2.12 & 3.41 & 6.54 & 7.88 & 1.88 & 65.81 & 29.97 & 0.43 & 6.17 & 1.78 & 248.1 \\
SigLIP2-B & 1.59 & 3.01 & 7.33 & 10.61 & 11.38 & 1.86 & 97.91 & 76.01 & 0.87 & 9.89 & 2.61 & 222.9 \\
PE-L & 1.72 & 2.80 & 5.91 & 10.90 & 9.88 & 2.86 & 90.07 & 56.80 & 0.88 & 8.19 & 2.84 & 221.5 \\
LangPE-L & 1.48 & 2.99 & 6.27 & 8.87 & 10.24 & 2.12 & 108.83 & 61.86 & 0.68 & 9.03 & 2.46 & 196.7 \\
SpatialPE-L & 1.16 & 1.82 & 4.67 & 6.62 & 8.57 & 1.35 & 55.46 & 45.79 & 0.51 & 7.23 & 1.86 & 247.1 \\
\midrule
\multicolumn{13}{c}{\textbf{Latent-space (VAE)}} \\
\midrule
SD-VAE-EMA & 1.38 & 1.92 & 7.71 & 6.94 & 19.15 & 2.92 & 48.52 & 87.44 & 0.62 & 20.39 & 2.43 & 259.6 \\
SD-VAE-EMA + REG & 1.32 & 1.74 & 7.24 & 7.55 & 18.47 & 2.04 & 55.02 & 84.72 & 0.69 & 20.14 & 2.34 & 271.6 \\
SD-VAE-MSE & 1.45 & 2.15 & 7.61 & 7.82 & 21.82 & 3.10 & 55.66 & 86.97 & 0.70 & 25.01 & 2.56 & 259.7 \\
SDXL-VAE & 1.69 & 2.74 & 9.26 & 8.64 & 21.55 & 3.38 & 70.85 & 109.22 & 0.75 & 22.66 & 3.07 & 256.0 \\
SD3.5-VAE & 1.51 & 2.14 & 7.76 & 6.19 & 15.16 & 3.04 & 75.53 & 89.20 & 0.55 & 14.04 & 2.64 & 262.9 \\
FLUX.1-VAE & 2.04 & 3.89 & 9.25 & 8.19 & 19.54 & 3.37 & 107.70 & 105.18 & 0.82 & 18.62 & 3.55 & 245.7 \\
FLUX.2-VAE & 0.89 & 1.07 & 4.32 & 3.90 & 10.75 & 0.90 & 24.98 & 43.59 & 0.31 & 9.84 & 1.37 & 272.7 \\
FLUX.2-VAE + REG & 0.92 & 0.95 & 4.27 & 3.91 & 10.17 & 1.06 & 37.56 & 42.47 & 0.31 & 9.42 & 1.44 & 294.1 \\
Qwen-Image-VAE & 1.85 & 4.56 & 9.77 & 9.19 & 25.46 & 2.75 & 159.26 & 118.42 & 0.90 & 25.97 & 3.01 & 238.9 \\
E2E-VAVAE & 1.08 & 1.99 & 4.51 & 4.71 & 9.83 & 2.13 & 62.17 & 50.75 & 0.36 & 9.17 & 1.65 & 275.4 \\
E2E-FLUX.1-VAE & 1.07 & 1.83 & 5.12 & 4.68 & 11.76 & 1.16 & 50.73 & 53.43 & 0.36 & 10.75 & 1.67 & 266.3 \\
E2E-SD3.5-VAE & 1.16 & 1.30 & 4.57 & 5.49 & 11.12 & 1.10 & 42.14 & 48.18 & 0.42 & 10.25 & 1.62 & 265.4 \\
E2E-Qwen-Image-VAE & 1.06 & 2.26 & 4.57 & 4.63 & 11.33 & 1.54 & 61.19 & 48.81 & 0.37 & 10.24 & 1.55 & 261.4 \\
\midrule
\multicolumn{13}{c}{\textbf{Pixel-space}} \\
\midrule
JiT & 2.38 & 4.59 & 9.57 & 13.70 & 22.51 & 3.97 & 146.37 & 113.63 & 1.23 & 21.21 & 4.08 & 231.2 \\
PixNerd & 2.45 & 4.01 & 8.33 & 11.67 & 20.65 & 4.24 & 104.21 & 86.10 & 0.96 & 18.94 & 4.17 & 213.8 \\
PixelGen & 2.26 & 4.34 & 7.80 & 13.14 & 17.97 & 3.96 & 138.33 & 86.50 & 1.12 & 15.73 & 3.97 & 247.4 \\
\midrule
\multicolumn{13}{c}{\textbf{One-/Few-step}} \\
\midrule
MeanFlow (SD-VAE-MSE, NFE=1) & 3.71 & 4.71 & 17.35 & 15.54 & 43.69 & 6.19 & 116.13 & 203.45 & 1.35 & 49.72 & 6.60 & 206.7 \\
MeanFlow (SD-VAE-MSE, NFE=2) & 3.19 & 3.25 & 13.19 & 9.84 & 28.42 & 7.93 & 69.05 & 148.75 & 0.83 & 27.45 & 5.40 & 226.5 \\
\bottomrule
\end{tabular}
\caption{\textbf{Systematic comparison on ImageNet-256 with CFG.} A single \nanogen{} backbone and training recipe applied across diverse latent-space tokenizers and pixel-space architectures. All models are trained for 80 epochs with $\sim$615M parameters and reported with classifier-free guidance applied at applied at the best CFG scale of each method, selected per method via a sweep over the guidance interval fixed at $[0.0, 0.9]$. FDr and MIND are computed against five vision encoders: Inception, ConvNeXt, DINOv2, MAE, and SigLIP. Except for the REPA-E family where the E2E VAEs are frozen after end-to-end training, REPA is not used.}
\label{tab:in1k_systematical_cfg}
\end{table*}

\textbf{First}, the best FID=1.37 is achieved by FLUX.2-VAE~\citep{flux2}, followed by DiTs trained with the REPA-E VAE family~\citep{repae}, with FID around 1.5 and 1.6. It is unclear how the FLUX.2-VAE is trained, but its architecture shares the same batch normalization layer as in REPA-E~\citep{repae} design, so perhaps they share similar mechanisms of end-to-end VAE and DiT tuning. \textbf{Second}, the RAE family has slight higher FID, with the better ones around 1.7-1.9. DINOv3-B has the best FID=1.74 among RAEs, while DINOv2-B has an FID = 1.96. Comparing results in Table \ref{tab:in1k_systematical_cfg} with Table \ref{tab:in1k_systematical}, it remains unclear how RAE benefits further from CFG, an open direction we leave to future work. \textbf{Third}, traditional VAEs such as SD-VAE~\citep{ldm} and SD3.5-VAE~\citep{sd3} lag behind. That said, we note that at 80 epochs the performance gap is largely driven by convergence speed, which is accelerated by well-structured latents such as those of RAE and REPA-E; we expect the gap relative to standard VAEs to narrow with longer training.

\textbf{Pixel-space generation and MeanFlow.}
For pixel-space generation, \nanogen operates directly on pixels without any latent tokenizer. In Tab.~\ref{tab:in1k_systematical_cfg}, we train PixNerd~\citep{pixnerd}, JiT~\citep{jit}, and PixelGen~\citep{pixelgen} using \nanogen. We observe that pixel-space FID is typically higher than latent-space FID at 80 training epochs. Similar to traditional VAE methods, the evaluated pixel-space methods do not show accelerated convergence at 80 epochs.

\nanogen also supports one-/few-step generation. We train MeanFlow~\citep{meanflow} on the SD-VAE-MSE~\citep{stabilityai2025sdvae} latent, and evaluate the model for one or two inference steps. MeanFlow reaches FID $6.60$ and $5.40$ with one or two steps, respectively. Even so, MeanFlow still lags behind the multi-step methods on either latent-space or pixel-space generation.

In general, observations \textit{w.r.t} RAE, latent-space methods, pixel-space methods, and MeanFlow from Tab.~\ref{tab:in1k_systematical_cfg} align with our general impression. 


\begin{table}[t]
    \centering
    \scriptsize
    \setlength{\tabcolsep}{4mm}{
    \begin{tabular}{lccccc}
    \toprule
    \textbf{Method} & \textbf{Iters} & \textbf{\#Params} & \textbf{GenEval}$\uparrow$ & \textbf{DPG-Bench}$\uparrow$ & \textbf{GenAIBench}$\uparrow$\\
    \midrule
    \multicolumn{6}{c}{\textbf{Public models}} \\
    \midrule
    SD-3.5-Large & - & 8B & 0.691 & 0.842 & 0.767 \\
    FLUX-1 & - & 12B & 0.654 & 0.838 & 0.748 \\
    FLUX-2 & - & 32B & 0.854 & 0.870 & 0.841 \\
    Qwen-Image & - & 20B & 0.848 & 0.888 & 0.803 \\
    Z-Image-Turbo & - & 6B & 0.736 & 0.847 & 0.759 \\
    \midrule
    \multicolumn{6}{c}{\textbf{Latent-space (RAE)~\citep{rae}}} \\
    \midrule
    DINOv2-B & 100K & 615M & 0.628 & 0.810 & 0.707 \\
    DINOv2-B + REG & 100K & 619M & 0.608 & 0.808 & 0.702 \\
    DINOv3-B & 100K & 615M & 0.636 & 0.828 & 0.718 \\
    DINOv3-B + REG & 100K & 619M & 0.642 & 0.827 & 0.730 \\
    SigLIP2-B & 100K & 615M & 0.606 & 0.809 & 0.718 \\
    PE-L & 100K & 617M & 0.586 & 0.818 & 0.723 \\
    SpatialPE-L & 100K & 617M & 0.535 & 0.790 & 0.694 \\
    LangPE-L & 100K & 617M & 0.633 & 0.826 & 0.724 \\
    LangPE-L & 200K & 617M & 0.635 & 0.824 & 0.715 \\
    \midrule
    \multicolumn{6}{c}{\textbf{Latent-space (VAE)}} \\
    \midrule
    SD-VAE-EMA & 100K & 611M & 0.578 & 0.804 & 0.691 \\
    SD-VAE-EMA + REG & 100K & 615M & 0.570 & 0.792 & 0.691 \\
    SD-VAE-MSE & 100K & 611M & 0.624 & 0.813 & 0.701 \\
    SDXL-VAE & 100K & 611M & 0.617 & 0.812 & 0.705 \\
    SD3.5-VAE & 100K & 612M & 0.640 & 0.818 & 0.702 \\
    Qwen-Image-VAE & 100K & 612M & 0.611 & 0.802 & 0.704 \\
    E2E-FLUX.1-VAE & 100K & 612M & 0.625 & 0.823 & 0.706 \\
    E2E-SD3.5-VAE & 100K & 612M & 0.637 & 0.840 & 0.715 \\
    E2E-Qwen-Image-VAE & 100K & 612M & 0.691 & 0.835 & 0.714 \\
    FLUX.1-VAE & 100K & 612M & 0.559 & 0.796 & 0.684 \\
    FLUX.1-VAE & 200K & 612M & 0.544 & 0.816 & 0.687 \\
    FLUX.2-VAE & 100K & 612M & 0.675 & 0.830 & 0.712 \\
    FLUX.2-VAE & 200K & 612M & 0.625 & 0.841 & 0.713 \\
    FLUX.2-VAE + REG & 100K & 616M & 0.687 & 0.830 & 0.722 \\
    E2E-VAVAE & 100K & 611M & 0.632 & 0.824 & 0.703 \\
    E2E-VAVAE & 200K & 611M & 0.679 & 0.836 & 0.716 \\
    \midrule
    \multicolumn{6}{c}{\textbf{Pixel-space}} \\
    \midrule
    JiT & 100K & 615M & 0.516 & 0.782 & 0.674 \\
    PixNerd & 100K & 615M & 0.484 & 0.777 & 0.643 \\
    PixelGen & 100K & 615M & 0.554 & 0.798 & 0.678 \\
    \midrule
    \multicolumn{6}{c}{\textbf{One-/Few-step}} \\
    \midrule
    MeanFlow (SD-VAE-MSE, NFE=1) & 100K & 613M & 0.287 & 0.688 & 0.582 \\
    MeanFlow (SD-VAE-MSE, NFE=2) & 100K & 613M & 0.341 & 0.721 & 0.602 \\
    \bottomrule
    \end{tabular}
    }
    \caption{\textbf{DiT comparison on text-to-image generation.} We use \nanogen{} for training DiTs across the latent space, pixel space, and MeanFlow under a unified backbone and training recipe. We report GenEval~\citep{geneval}, DPG-Bench~\citep{dpgbench}, and GenAIBench~\citep{genaibench}. For reference, we additionally report the benchmark results of several public T2I models, such as SD3.5-Large~\citep{sd3}, FLUX-1~\citep{flux}, FLUX-2~\citep{flux2}, Qwen-Image~\citep{qwenimage}, and Z-Image-Turbo~\citep{zimage}.
    }
    \label{tab:t2i_systematical}
\end{table}

\subsection{Benchmarking T2I Generation Methods}
\label{sec:t2i_results}

We take the same diffusion methods evaluated on ImageNet (Sec.~\ref{sec:imagenet}) and re-train each into a T2I model under a common protocol. For each method we report results using T2I metrics including GenEval~\citep{geneval}, DPG-Bench~\citep{dpgbench}, and GenAIBench~\citep{genaibench}. All variants follow the training recipe of Sec.~\ref{sec:t2i_impl}. Tab.~\ref{tab:t2i_systematical} summarises the systematic comparison of all methods under the latent-space and pixel-space setups. We have five major observations. 



\begin{figure*}[t]
  \centering
  \includegraphics[width=\textwidth]{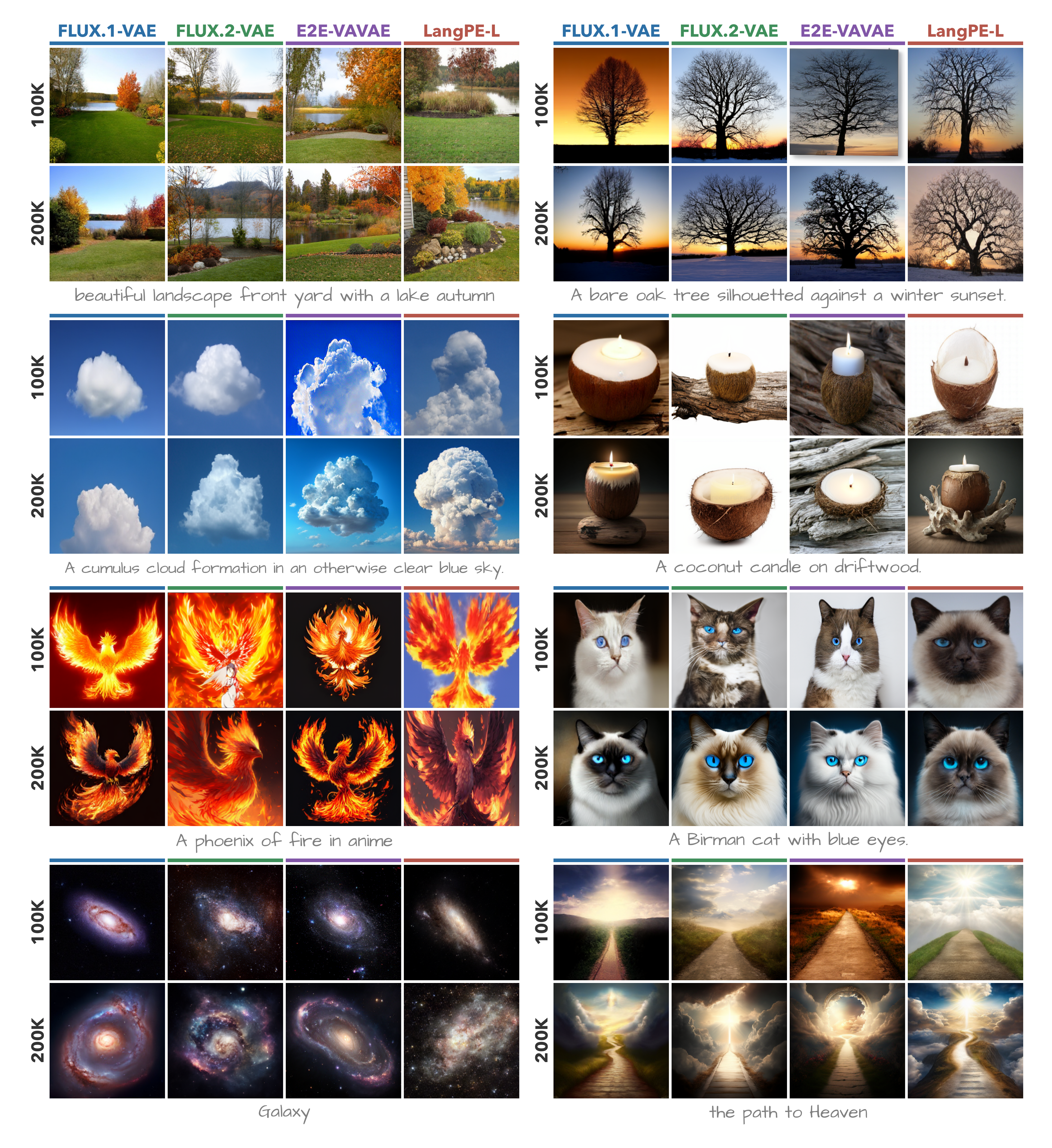}
  \vspace{-0.8cm}
  \caption{\textbf{Text-to-image qualitative samples at $256{\times}256$.} Curated qualitative samples from \nanogen latent-space methods trained for $100\text{K}$ and $200\text{K}$ iterations at batch size $1024$, evaluated on a shared set of text prompts. Quantitative scores for the same methods are reported in Tab.~\ref{tab:t2i_systematical}.}
  \label{fig:t2i_qualitative}
\end{figure*}

\textbf{First}, in the state-of-the-art frontier, ImageNet ranking does not robustly predict the T2I ranking. For example, RAE with SpatialPE-L has very good ImageNet FID, but its T2I performance is among the worst across various metrics. \textbf{Second}, different metrics to some extent disagree with each other. For example, E2E-Qwen-Image-VAE is one of the strongest if we look at GenEval and DPG-Bench metrics but it falls into the second tier under the GenAIBench metric.  
\textbf{Third}, the class-conditional ImageNet trend is consistent with the T2I-metric trend if we look at broader method category ranking. That is, improved latent-space methods (RAE, FLUX.2-VAE, and REPA-E) \textgreater\ traditional latent-space methods \textgreater\ pixel-space methods \textgreater\ MeanFlow (Tab.~\ref{tab:in1k_systematical_cfg}, Tab.~\ref{tab:t2i_systematical}, and Fig.~\ref{fig:fid_correlation_withpixel_appendix} [a,b]). From this perspective, ImageNet signals are useful. End-to-end VAE tuning improves both ImageNet FID and T2I metrics, including FLUX.1-VAE and Qwen-Image-VAE. But a better ImageNet FID does not predict a better T2I score across different methods, which remains the case without CFG (Fig.~\ref{fig:fid_correlation_nocfg_appendix}). Most state-of-the-art methods report FID between 1 and 2, which fall into the most uncorrelated regions (Fig.~\ref{fig:fid_correlation_withpixel_appendix}b). \textbf{Fourth}, comparing numbers in Table \ref{tab:t2i_systematical} with those of public T2I models in the same table, the pre-trained T2I models using \nanogen are generally worse, which is consistent with our perception. \textbf{Last}, when we train T2I for 200K steps, the performance generally remains similar or improves slightly under the three metrics. This observation is interesting: upon visual check in Fig.~\ref{fig:t2i_qualitative}, images at 200K training are better than those at 100K. We suspect that better metrics should be proposed.

\begin{figure*}[t]
  \centering
  \includegraphics[width=\textwidth]{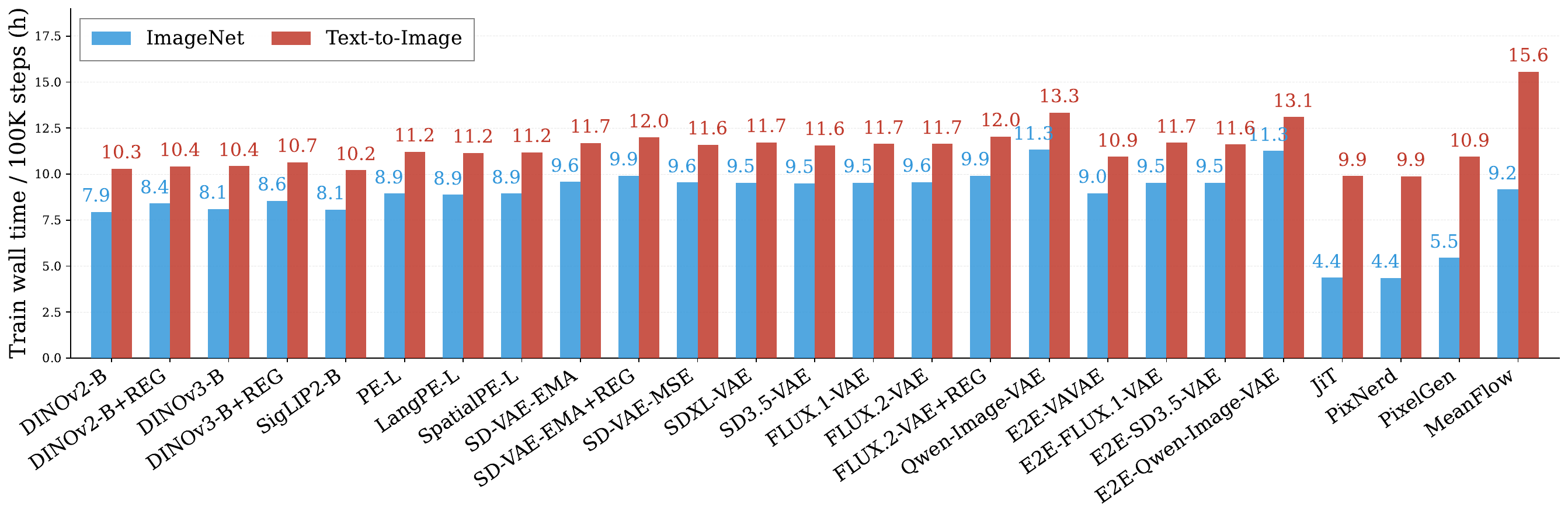}
  \caption{\textbf{Wall-clock training time comparison of ImageNet and T2I setups.} We record time for $100\text{K}$ steps for 25 DiT methods. We use 32 H200 GPUs with a unified training recipe in \nanogen.
  Training T2I remains efficient across all methods. Moreover, training cost is comparable across latent-space methods, while pixel-space methods such as JiT~\citep{jit}, PixNerd~\citep{pixnerd}, and PixelGen~\citep{pixelgen}, are much cheaper to train on ImageNet because they do not compute latents from VAEs. RAE~\citep{rae} methods are marginally faster to train than VAE methods, because RAE relies on transformer-based vision encoders whereas VAEs mainly use a convolution-based U-Net structure. MeanFlow~\citep{meanflow} is much slower than other T2I methods, as it computes the MeanFlow objective with \texttt{torch.jvp}, which adds substantial computational overhead. If not specified, ImageNet and T2I models are trained with 100K steps in this paper.}
  \label{fig:training_time}
\end{figure*}

\textbf{Are the T2I results competitive?} As shown in Fig.~\ref{fig:training_time}, we use a small compute budget around 10 hours of wall-clock time with 32 H200 GPUs, though all setups are runnable on 8 H200 GPUs. RAEv2 \citep{raev2} reports a GenEval score of 0.624 using a SigLIP2-B encoder and an 875M diffusion model, pre-trained on the same dataset for 150K iterations; in comparison, we report a GenEval score of 0.691 using E2E-Qwen-Image-VAE and 0.633 using RAE based on LangPE-L. We did not find many other public numbers of pre-trained-only T2I models under similar compute. On the other hand, many papers report T2I models after supervised fine-tuning on the BLIP-3o-60K~\citep{blip3o} dataset, where their GenEval scores are around 0.85-0.90. We did similar fine-tuning on top of our FLUX.1-VAE T2I checkpoint and can achieve 0.90 GenEval score. While GenEval scores can be very high, this might be due to metric hacking, and the resulting models may not be universally better. We call on more hack resistant evaluation for T2I models.

\Paragraph{Recommended usage.} Our recommendation is that future DiT papers report \diffbench, which includes both ImageNet and T2I generation, rather than any single axis. Methods that improve \diffbench are more likely to reflect broadly useful progress; methods that improve one axis but regress another may still be valuable, but should be labelled as task-specific improvements rather than general DiT advances.

\section{Background and Related Work}
\label{sec:background}

\Paragraph{Diffusion and flow matching.}
Diffusion models~\citep{ddpm, song-diffusion, scorematching} have become the dominant framework for visual generation.
They corrupt data with Gaussian noise along a forward path $x_t = \alpha_t x_0 + \sigma_t \epsilon$ and train a network predicting the noise $\epsilon$ to reverse it.
Flow matching~\citep{fmgen, fm} instead regresses the velocity field along the same path.
The prediction target (the noise $\epsilon$, the clean signal $x_0$, or the velocity $v$) is interchangeable up to a time-dependent reweighting~\citep{edm, salimans2022progressive, sct}. We primarily adopt $v$-prediction (Sec.~\ref{sec:method}).
At inference, a sample is generated by solving the ordinary differential equation (ODE) defined by the learned velocity field, integrating from noise to data.

\Paragraph{Diffusion transformers} (DiTs)~\citep{dit} replace the U-Net backbone of earlier diffusion models with a transformer over image patches.  
Later works have refined it, including SiT~\citep{sit}, long skip connections \citep{uvit}, joint text--image attention in MMDiT~\citep{sd3}, and encoder--decoder split in DDT~\citep{ddt}. \nanogen currently is built on the DDT backbone (Sec.~\ref{sec:method}). 
Some other works remove the latent tokenizer and train diffusion transformers directly in pixel space, including PixNerd~\citep{pixnerd}, JiT~\citep{jit}, and PixelGen~\citep{pixelgen}.
Recent works also train text-to-image models with simple and fully open recipes, including i1~\citep{i1} and MiniT2I~\citep{minit2i}. These works focus on ablating training recipes for a single strong T2I model. In contrast, we fix a unified recipe and compare a wide range of recent diffusion methods in a fair setup, with minimal engineering effort and compute cost. We do not aim for achieving state-of-the-art performance on T2I tasks.

\Paragraph{Tokenizers and representation alignment.}
Latent diffusion models~\citep{ldm} diffuse in the lower-dimensional latent space of a pretrained VAE~\citep{vae}, which makes high-resolution training computationally tractable.
Pretrained vision-encoder representations have been shown to substantially accelerate and improve diffusion training:
representation alignment (REPA)~\citep{repa} aims to align the internal features of DiTs with those of a frozen vision encoder; REPA-E~\citep{repae} uses this alignment signal to additionally tune the VAE end to end.
Representation autoencoders (RAE)~\citep{rae, raev2} merge their ideas and use a frozen vision encoder directly as tokenizer, such as DINOv2~\citep{dinov2}, DINOv3~\citep{dinov3}, SigLIP2~\citep{siglip2}, or the Perception Encoder~\citep{perceptionencoder}, so a single representation provides both compression and semantic structure. 
Recent analysis studies which properties of these encoders benefit generation most~\citep{irepa}.

\Paragraph{ImageNet metrics and their limitation.}
Generation quality on ImageNet is typically measured by Frechet Inception Distance (FID)~\citep{fid}, Inception Score~\citep{is}, and precision/recall~\citep{precrecall}. FID has limitations: it is sensitive to image resizing and preprocessing, depends on a frozen Inception-v3~\citep{inceptionv3} classifier trained on a related domain, and has now saturated. 
Improvemet over FID includes sFID~\citep{sfid}, FDr~\citep{fdloss}, and MIND~\citep{mind}. 

\Paragraph{Text-to-image evaluation.}
Early text-to-image evaluation uses FID on MS-COCO~\citep{coco} for image fidelity and diversity. They also use CLIPScore~\citep{clip} to evaluate prompt alignment. These metrics are coarse and do not accurately reflect model quality.
Metrics with more accurate measurement of prompt alignment includes reward models such as ImageReward~\citep{imagereward}, HPSv2~\citep{hpsv2}, and PickScore~\citep{pickscore}, compositional benchmarks such as GenEval~\citep{geneval} and DPGBench~\citep{dpgbench}, and VLM-based scorers such as VQAScore~\citep{vqascore}, UnifiedReward~\citep{unified-reward}, and Qwen-Image-Bench~\citep{qwen-image-bench}.
These metrics aim to fully reflect human preference. The latter is currently best captured by large-scale human-labeled arenas such as the Artificial Analysis Arena~\citep{artificialanalysis}. It aggregates human pairwise votes into ELO ratings.


\Paragraph{Holistic benchmarking.}
As fields mature, single dimensional leaderboards tend to give way to multi-task evaluation.
In language modeling, holistic leaderboards such as HELM~\citep{helm} and BIG-Bench~\citep{big-bench} have replaced single-axis rankings. In image generation, HEIM \citep{heim} is an \textit{evaluation} platform for \textit{already trained} T2I models. In fact, the development of DiTs requires \textit{training and evaluation} of the same DiT idea on \textit{both ImageNet and T2I tasks}, which no current infrastructure can support. 

\section{Conclusion}
\label{sec:conclusion}
Diffusion transformer research has matured to the point where single-benchmark evaluation is no longer enough. In this paper we introduce \nanogen, a training and evaluation framework that removes the engineering barrier to training and evaluating DiT methods on the T2I task, and use it to show that ImageNet rankings do not reliably predict text-to-image performance.
Finally, we package the two evaluation axes: ImageNet and T2I generation, into \diffbench and argue for its adoption as the default DiT benchmark. Our hope is that making holistic evaluation cheap, both engineering-wise and computationally, will shift the field toward progress that is broad rather than local.



\Paragraph{What we are not claiming.} We do not claim that ImageNet and FID are no longer useful; they are still a good platform for generative modeling. 
Besides, we do not claim that \diffbench is a permanent benchmark. It should be refreshed as methods begin to saturate it.

\Paragraph{Limitations.} First, the ImageNet-T2I correlation (Fig.~\ref{fig:fid_correlation}) is measured at the scale and compute we could afford and may look different at other scales.
Second, the benchmarking results (Tab.~\ref{tab:in1k_systematical_cfg} and Tab.~\ref{tab:t2i_systematical}) are obtained after 100K iterations with a batch size of 1,024, which we can afford. Longer training will improve the generated image quality (see Fig.~\ref{fig:t2i_qualitative}).


\Paragraph{Future work.} There are several promising directions. First, \diffbench can be broadened to other generative modalities such as world models, videos, and 3D, so that cross-task evaluation captures the full scope of diffusion-based generation.
Second, current T2I metrics could be hacked through fine-tuning on a curated dataset, so better hack-resistant mechanisms are needed for T2I metrics.
Finally, we envisage \diffbench as a living and community-maintained leaderboard that is periodically refreshed to keep pace with advances in methodology.


\section{Contributors}
\label{sec:contributors}

\textbf{Code Development.} Jaskirat Singh led most of the development for the unified codebase. Xingjian Leng added online T2I evaluation suites, REG and  pixel-space methods. The individual contributions are:

\begin{itemize}
    \item \emph{Jaskirat Singh.} Added stage1 (VAE/RAE) and stage2 (diffusion model) training across different tasks (ImageNet, T2I), 80+ different vision encoders for RAE and VAE, autoguidance (RAE), REPA, unified dataloader, in-context conditioning, MeanFlow, Gmuon optimiser, online gFID/rFID evaluation, simple T2I (using 256 text embedding tokens for T2I instead of 8 class condition tokens in ImageNet).
    \item \emph{Xingjian Leng.} Helped add online evaluation for T2I experiments (GenEval, DPG-Bench, and GenAIBench). He also added REG, pixel-space, and MeanFlow implementation and  ran final experiments/results reported in the paper.
\end{itemize}

\textbf{Paper Writing and Advising.} Jaskirat Singh wrote the initial draft of the full paper. Xingjian Leng and Zhanhao Liang helped with most of the paper writing and final draft. Liang Zheng advised the project and paper writing. Ethan Smith, Martin Bell, Aninda Saha, and Yuhui Yuan provided feedback on project and paper writing.

\clearpage
{
    \small
    \bibliographystyle{iclr2025_conference}
    \bibliography{main}
}

\clearpage
\appendix
\section{Additional results}
\label{sec:appendix}

\subsection{ImageNet systematic comparison without CFG}
\label{sec:appendix_in1k_nocfg}

\begin{table*}[h]
\centering
\scriptsize
\setlength{\tabcolsep}{0.45mm}
\begin{tabular}{l|ccccc|ccccc|cc}
\toprule
\multirow{2}{*}{\textbf{Method}} & \multicolumn{5}{c|}{\textbf{FDr}$\downarrow$} & \multicolumn{5}{c|}{\textbf{MIND}$\downarrow$} & \multirow{2}{*}{\textbf{FID}$\downarrow$} & \multirow{2}{*}{\textbf{IS}$\uparrow$} \\
\cmidrule(lr){2-6} \cmidrule(lr){7-11}
 & \textbf{Incep.} & \textbf{ConvNeXt} & \textbf{DINOv2} & \textbf{MAE} & \textbf{SigLIP} & \textbf{Incep.} & \textbf{ConvNeXt} & \textbf{DINOv2} & \textbf{MAE} & \textbf{SigLIP} & & \\
\midrule
\multicolumn{13}{c}{\textbf{Latent-space (RAE)~\citep{rae}}} \\
\midrule
DINOv2-B & 1.32 & 2.33 & 3.14 & 6.33 & 7.56 & 2.54 & 70.53 & 26.52 & 0.44 & 6.30 & 2.14 & 211.9 \\
DINOv2-B + REG & 1.29 & 2.45 & 3.26 & 6.42 & 7.82 & 2.23 & 76.20 & 27.98 & 0.45 & 6.47 & 2.08 & 207.7 \\
DINOv3-B & 1.33 & 2.82 & 3.89 & 6.80 & 8.17 & 2.16 & 97.15 & 33.62 & 0.45 & 6.45 & 2.15 & 200.8 \\
DINOv3-B + REG & 1.32 & 2.71 & 3.79 & 6.71 & 8.11 & 2.35 & 90.39 & 32.33 & 0.45 & 6.38 & 2.15 & 204.4 \\
SigLIP2-B & 2.11 & 4.20 & 8.55 & 10.97 & 11.76 & 2.45 & 162.71 & 85.81 & 0.90 & 10.15 & 3.48 & 179.4 \\
PE-L & 1.86 & 3.15 & 6.17 & 11.01 & 9.87 & 3.05 & 97.69 & 58.40 & 0.89 & 8.08 & 3.08 & 206.6 \\
LangPE-L & 1.65 & 3.48 & 6.66 & 8.97 & 10.28 & 2.53 & 138.39 & 64.01 & 0.69 & 9.07 & 2.76 & 182.2 \\
SpatialPE-L & 2.18 & 4.17 & 7.30 & 7.01 & 10.74 & 2.35 & 206.52 & 69.92 & 0.54 & 9.16 & 3.61 & 160.4 \\
\midrule
\multicolumn{13}{c}{\textbf{Latent-space (VAE)}} \\
\midrule
SD-VAE-EMA & 5.97 & 10.14 & 18.17 & 11.11 & 35.43 & 9.56 & 688.82 & 245.95 & 0.99 & 38.87 & 10.16 & 113.4 \\
SD-VAE-EMA + REG & 3.14 & 5.14 & 11.30 & 9.58 & 25.63 & 3.45 & 246.05 & 145.13 & 0.87 & 28.41 & 5.39 & 162.7 \\
SD-VAE-MSE & 5.97 & 10.36 & 17.84 & 12.03 & 38.19 & 9.34 & 676.48 & 243.23 & 1.07 & 43.99 & 10.15 & 112.4 \\
SDXL-VAE & 7.50 & 12.58 & 21.50 & 13.57 & 40.39 & 12.96 & 858.54 & 297.59 & 1.21 & 44.20 & 12.88 & 104.7 \\
SD3.5-VAE & 5.96 & 10.21 & 17.97 & 10.39 & 30.85 & 8.50 & 634.85 & 240.52 & 0.96 & 30.70 & 10.18 & 111.7 \\
FLUX.1-VAE & 9.28 & 16.66 & 23.27 & 14.02 & 41.46 & 14.81 & 1072.69 & 316.55 & 1.41 & 42.58 & 15.75 & 86.6 \\
FLUX.2-VAE & 2.76 & 4.65 & 8.56 & 5.15 & 17.41 & 2.53 & 234.79 & 96.75 & 0.41 & 16.46 & 4.53 & 146.9 \\
FLUX.2-VAE + REG & 2.56 & 4.06 & 7.91 & 5.17 & 16.74 & 2.11 & 183.74 & 88.65 & 0.41 & 15.82 & 4.19 & 155.8 \\
Qwen-Image-VAE & 6.52 & 13.25 & 19.78 & 13.45 & 41.51 & 9.34 & 804.53 & 270.50 & 1.31 & 44.01 & 10.86 & 108.9 \\
E2E-VAVAE & 2.64 & 5.90 & 8.59 & 6.33 & 16.30 & 2.20 & 277.26 & 103.04 & 0.51 & 15.54 & 4.27 & 147.6 \\
E2E-FLUX.1-VAE & 3.83 & 6.89 & 10.92 & 6.67 & 21.28 & 4.23 & 373.50 & 129.48 & 0.53 & 20.39 & 6.30 & 134.3 \\
E2E-SD3.5-VAE & 3.37 & 5.10 & 9.21 & 7.07 & 18.70 & 3.19 & 266.87 & 108.44 & 0.55 & 17.69 & 5.32 & 140.8 \\
E2E-Qwen-Image-VAE & 3.11 & 6.80 & 9.57 & 6.46 & 19.85 & 2.84 & 337.95 & 113.68 & 0.53 & 18.88 & 4.98 & 138.4 \\
\midrule
\multicolumn{13}{c}{\textbf{Pixel-space}} \\
\midrule
JiT & 12.82 & 22.87 & 28.16 & 23.59 & 53.76 & 24.19 & 1632.56 & 412.33 & 2.15 & 55.40 & 21.72 & 65.0 \\
PixNerd & 12.18 & 21.42 & 25.08 & 19.67 & 48.24 & 22.77 & 1581.55 & 345.19 & 1.71 & 49.33 & 20.61 & 63.9 \\
PixelGen & 7.01 & 13.93 & 17.89 & 18.98 & 34.49 & 9.07 & 769.26 & 222.02 & 1.68 & 32.35 & 12.10 & 104.0 \\
\midrule
\multicolumn{13}{c}{\textbf{One-/Few-step}} \\
\midrule
MeanFlow (SD-VAE-MSE, NFE=1) & 14.56 & 24.09 & 37.39 & 21.15 & 70.61 & 31.56 & 1993.71 & 571.19 & 1.88 & 81.85 & 24.83 & 61.4 \\
MeanFlow (SD-VAE-MSE, NFE=2) & 12.24 & 21.97 & 34.62 & 16.23 & 58.31 & 24.85 & 1860.67 & 524.10 & 1.42 & 62.69 & 20.58 & 63.0 \\
\bottomrule
\end{tabular}
\caption{\textbf{Systematic comparison on ImageNet-256 without CFG.} All training setup is kept the same as Tab.~\ref{tab:in1k_systematical_cfg}, but evaluated without classifier-free guidance.}
\label{tab:in1k_systematical}
\end{table*}

\clearpage

\subsection{ImageNet-FID and T2I metrics correlation including pixel-space methods}
\label{sec:appendix_fid_t2i_correlation}

\begin{figure*}[h!]
  \centering
  \begin{subfigure}[b]{0.85\textwidth}
    \centering
    \includegraphics[width=\linewidth]{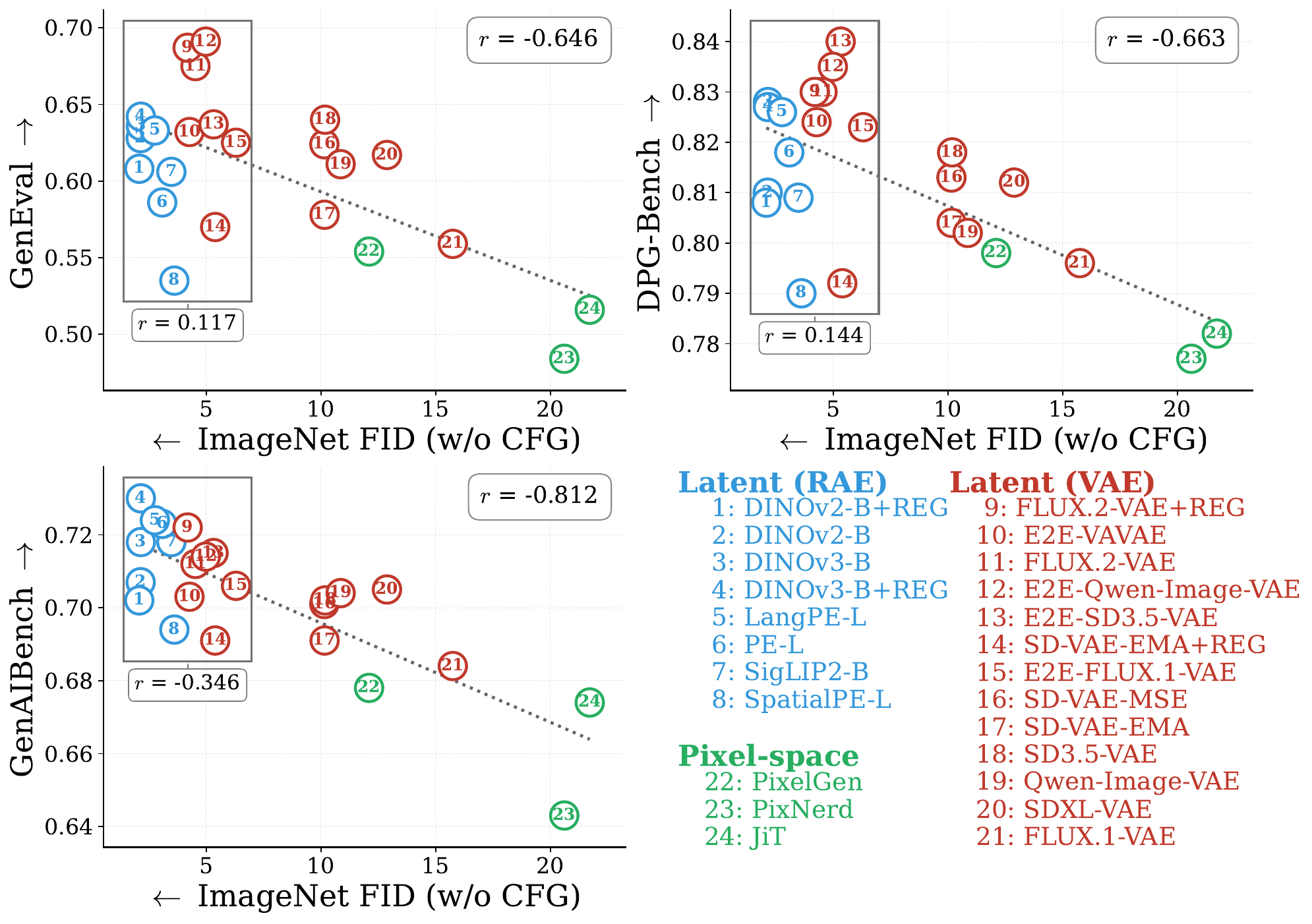}
    \vspace{-0.5cm}
    \caption{without CFG}
    \label{fig:fid_correlation_nocfg_appendix_withpixel}
  \end{subfigure}\hfill
  \begin{subfigure}[b]{0.85\textwidth}
    \centering
    \includegraphics[width=\linewidth]{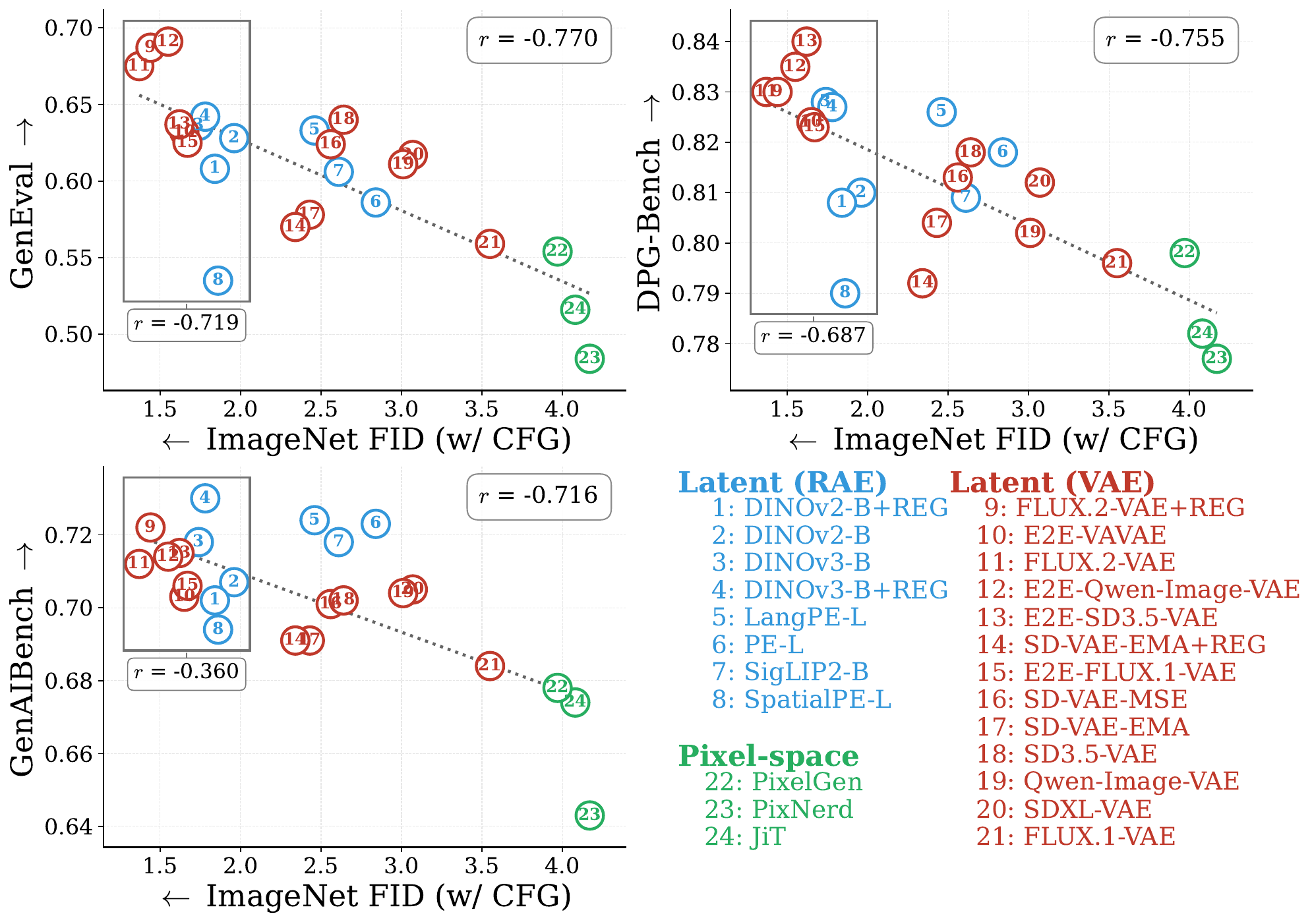}
    \vspace{-0.5cm}
    \caption{with CFG}
    \label{fig:fid_correlation_wcfg_appendix}
  \end{subfigure}
  \vspace{-0.25cm}
  \caption{\textbf{Correlation between ImageNet FID and T2I metrics including pixel-space methods.} Same setup as Fig.~\ref{fig:fid_correlation}, but with three pixel-space methods, JiT~\citep{jit}, PixNerd~\citep{pixnerd}, and PixelGen~\citep{pixelgen}, added. (a) Without classifier-free guidance and (b) under the best CFG scale of each method over a timestep interval of $[0.0, 0.9]$.}
  \label{fig:fid_correlation_withpixel_appendix}
  \vspace{-0.5cm}
\end{figure*}

Our main analysis excludes pixel-space methods. We include three pixel-space methods in Figure~\ref{fig:fid_correlation_withpixel_appendix}: JiT~\citep{jit}, PixNerd~\citep{pixnerd}, and PixelGen~\citep{pixelgen}, with both without and with CFG settings. Pixel-space methods are much worse than latent-space methods on both ImageNet FID and the T2I metrics, which would artificially raise the overall correlation. We therefore focus the main analysis on the latent-space frontier, where the correlation is not driven by these outliers. We also exclude MeanFlow methods, which would raise the correlation further.

\clearpage

\subsection{ImageNet-FID and T2I metrics correlation without CFG}
\label{sec:appendix_fid_t2i_correlation_nocfg}

\begin{figure*}[h!]
  \centering
  \includegraphics[width=0.975\textwidth]{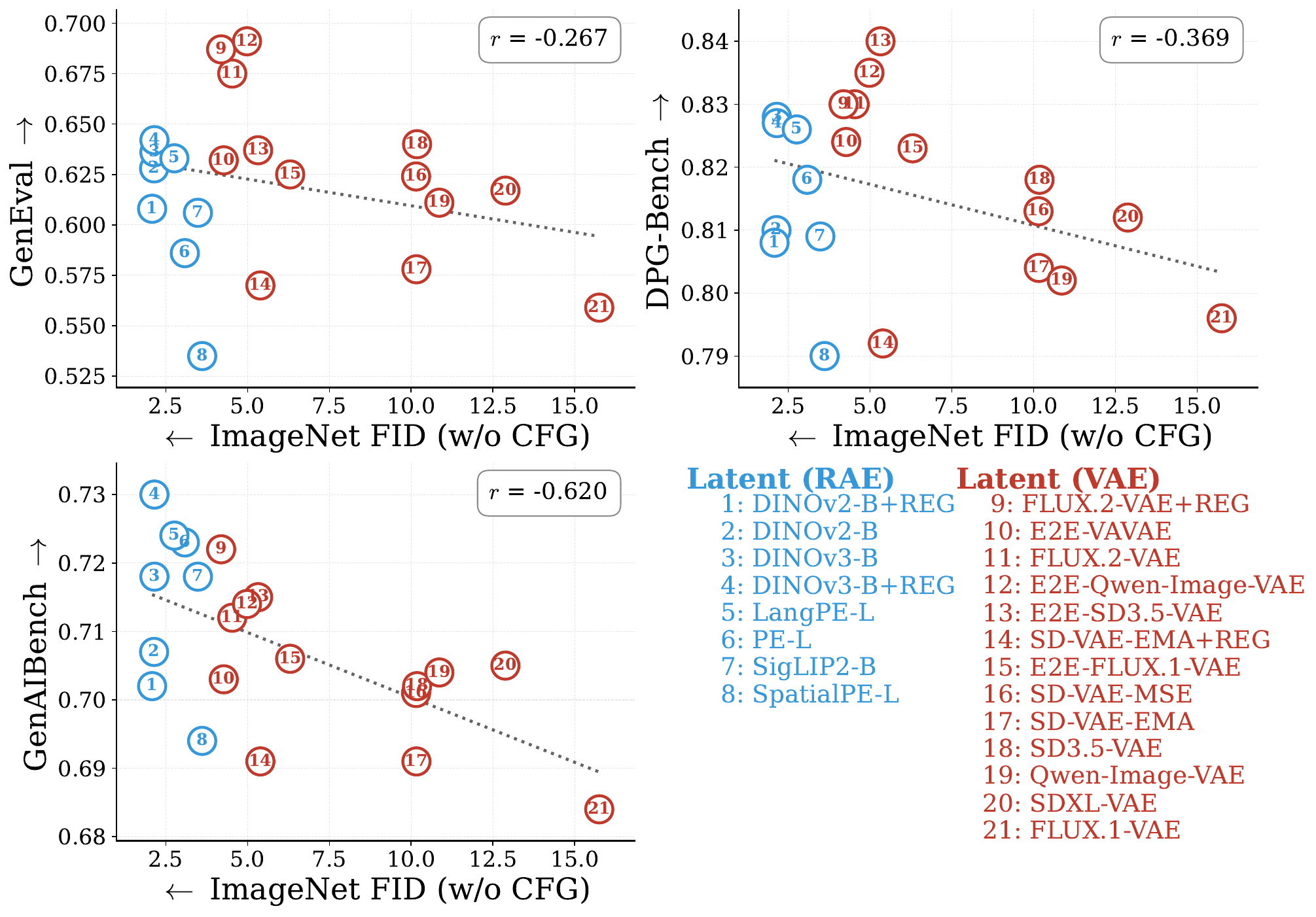}
  \caption{\textbf{Correlation between ImageNet FID and T2I metrics without CFG.} Same setup as Fig.~\ref{fig:fid_correlation}, but evaluated without classifier-free guidance.}
  \label{fig:fid_correlation_nocfg_appendix}
\end{figure*}

Our main correlation uses ImageNet FID with CFG, which matches the T2I protocol. We repeat the analysis in Figure~\ref{fig:fid_correlation_nocfg_appendix} with ImageNet FID evaluated without CFG. The Pearson $r$ values change across metrics, but there is still no evidence of a strong correlation between ImageNet FID and the T2I metrics. This finding does not depend on the CFG protocol used on ImageNet.

\end{document}